\documentclass{article} 

\usepackage{iclr2020_conference,times}
\usepackage{hyperref}
\usepackage{url}
\usepackage{graphicx}
\usepackage{algorithm}
\usepackage{algorithmic}

\usepackage{amsmath,amsfonts,bm}

\def\eqref#1{equation~\ref{#1}}

\def\1{\bm{1}}

\def\va{{\bm{a}}}
\def\vb{{\bm{b}}}

\def\vd{{\bm{d}}}

\def\vf{{\bm{f}}}

\def\vh{{\bm{h}}}

\def\vj{{\bm{j}}}

\def\vl{{\bm{l}}}
\def\vm{{\bm{m}}}
\def\vn{{\bm{n}}}
\def\vo{{\bm{o}}}
\def\vp{{\bm{p}}}
\def\vq{{\bm{q}}}

\def\vs{{\bm{s}}}

\def\vu{{\bm{u}}}
\def\vv{{\bm{v}}}
\def\vw{{\bm{w}}}

\def\vz{{\bm{z}}}

\def\mB{{\bm{B}}}

\def\mD{{\bm{D}}}

\def\mJ{{\bm{J}}}

\def\mL{{\bm{L}}}
\def\mM{{\bm{M}}}

\def\mP{{\bm{P}}}
\def\mQ{{\bm{Q}}}

\def\mS{{\bm{S}}}

\def\mU{{\bm{U}}}
\def\mV{{\bm{V}}}

\def\mZ{{\bm{Z}}}

\DeclareMathAlphabet{\mathsfit}{\encodingdefault}{\sfdefault}{m}{sl}
\SetMathAlphabet{\mathsfit}{bold}{\encodingdefault}{\sfdefault}{bx}{n}
\newcommand{\tens}[1]{\bm{\mathsfit{#1}}}
\def\tA{{\tens{A}}}
\def\tB{{\tens{B}}}

\def\tH{{\tens{H}}}

\def\tJ{{\tens{J}}}

\def\tL{{\tens{L}}}
\def\tM{{\tens{M}}}

\def\tO{{\tens{O}}}
\def\tP{{\tens{P}}}
\def\tQ{{\tens{Q}}}

\def\tU{{\tens{U}}}
\def\tV{{\tens{V}}}
\def\tW{{\tens{W}}}
\def\tX{{\tens{X}}}

\def\sB{{\mathbb{B}}}

\def\sD{{\mathbb{D}}}
\def\sF{{\mathbb{F}}}

\def\sJ{{\mathbb{J}}}

\def\sL{{\mathbb{L}}}
\def\sM{{\mathbb{M}}}

\def\sP{{\mathbb{P}}}
\def\sQ{{\mathbb{Q}}}

\def\sS{{\mathbb{S}}}

\def\sX{{\mathbb{X}}}

\newcommand{\R}{\mathbb{R}}

\title{A Parallel Implementation of Computing Mean Average Precision}

\author{Beinan Wang\\
\texttt{beinanwangcanada@gmail.com}}

\iclrfinalcopy
\begin{document}

\maketitle

\begin{abstract}
Mean Average Precision (mAP) has been widely used for evaluating the quality of object detectors, but an efficient implementation is still absent. Current implementations can only count true positives (TP's) and false positives (FP's) for one class at a time by looping through every detection of that class sequentially. Not only are these approaches inefficient, but they are also inconvenient for reporting validation mAP during training. We propose a parallelized alternative that can process mini-batches of detected bounding boxes (DTBB's) and ground truth bounding boxes (GTBB's) as inference goes such that mAP can be instantly calculated after inference is finished. Loops and control statements in sequential implementations are replaced with extensive uses of broadcasting, masking, and indexing. All operators involved are supported by popular machine learning frameworks such as PyTorch and TensorFlow. As a result, our implementation is much faster and can easily fit into typical training routines. A PyTorch version of our implementation is available at \url{https://github.com/bwangca/fast-map}.
\end{abstract}

\section{Introduction}\label{section_introduction}
In spite of the differences in determining TP's and FP's, prevalent object detection challenges such as Pascal VOC \citep{pascal_voc} and Microsoft COCO \citep{coco} adopt mAP as the standard evaluation metric. It is a common practice to minimize a set of surrogate losses instead of directly optimizing mAP because mAP is non-differentiable \citep{map_surrogate_loss}. However, changes in loss values may not correctly reflect changes in mAP. It may hence be more beneficial to save the best weights based on validation mAP, but this is rarely done in practice. Instead, the weights after a predetermined number of iterations are often chosen as the ultimate weights of an object detector. For example, the parameters of YOLOv3 \citep{yolov3} are finalized after feeding 500,200 batches of samples to the network.

The reason why many object detection algorithms do not use validation mAP as the criterion for selecting the best weights is seldom discussed. We find one possible cause is the lack of an efficient and convenient way to compute mAP. To the best of our knowledge, current implementations can only process one detection from one category at a time. The evaluation process is disjoint from the training loop and unsupported by hardware acceleration.

We present an alternative approach that handles a mini-batch of DTBB's and GTBB's at once. We utilize standard operators that are heavily optimized by many machine learning libraries, such as PyTorch \citep{pytorch} and TensorFlow \citep{tensorflow}. As a result, our implementation benefits immensely from hardware acceleration and can fit seamlessly into a training routine.
\section{Notation}\label{section_notation}
Before delving into mAP related content, we need to introduce the notation used throughout this paper. For expressing numbers and arrays, we follow \cite{deep_learning} as much as possible. Where it might be confusing to name a scalar with a single letter, we use plain text (e.g., mAP) instead.

When it comes to advanced indexing, such as selecting a subarray with an index array or a binary mask, we follow the convention of Numpy \citep{numpy}. Interested readers can find more details about Numpy in \cite{numpy_guide}.

Furthermore, there come times when a single "function" (e.g., \texttt{cumsum}) is more suitable in delivering the meaning of an otherwise extended expression. We are aware that mixing mathematical expressions and "code" may not be universally accepted. However, after weighing the pros and cons, we decide to combine the best from both to make algorithms concise.
\section{Review}\label{section_review}
For a dataset that contains $k$ classes of objects, the mAP of a detector at an intersection over union (IoU) threshold $t$ is defined in Equ. \ref{equation_map}.
\begin{equation}\label{equation_map}
\text{mAP@}t = \frac{1}{k}\Sigma_0^{k-1}\text{AP}\text{@}t
\end{equation}

\begin{figure}[tb]

\vskip 0.2in

\begin{center}

\centerline{\includegraphics[width=\columnwidth]{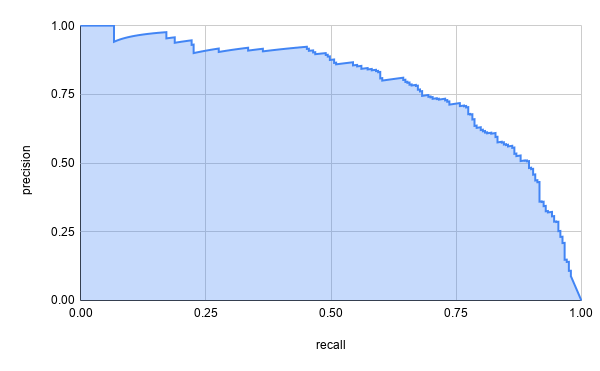}}

\caption{The "sofa" PR curve of Faster R-CNN on the Pascal VOC 2007 test set. The shaded area under the curve is equal to the AP, which ranges from 0 to 1.}

\label{figure_ap}

\end{center}

\vskip -0.2in

\end{figure}

The average precision (AP) for a particular class is the area under the precision-recall (PR) curve, as shown in Fig. \ref{figure_ap}. DTBB's have to be sorted by their confidence scores in descending order first so that the resulted PR curve is invariant to the order by which a detector receives images.
The recall at the $i$-th point (starting from 0) on the PR curve can be calculated according to Equ. \ref{equation_recall},
\begin{equation}\label{equation_recall}
q_i = \frac{1}{\hat{z}}\Sigma_0^i\1_\mathrm{\text{TP}}
\end{equation}
where $\hat{z}$ is the total number of "easy" GTBB's in the entire dataset for the class under investigation. There might be some "difficult" GTBB's that are either occluded or truncated by a large percentage. Not being able to detect them is not punished during evaluation.

On the other hand, precision does not measure the percentage of GTBB's detected. It instead emphasizes on the percentage of correct predictions made. The precision at the $i$-th point on the PR curve is defined in Equ. \ref{equation_precision}.
\begin{equation}\label{equation_precision}
p_i = \Sigma_0^i\frac{\1_\mathrm{\text{TP}}}{\1_\mathrm{\text{TP}} + \1_\mathrm{\text{FP}}}
\end{equation}
It is worth mentioning that the number of points on the PR curve is not necessarily equal to the number of DTBB's. Some DTBB's might match "difficult" GTBB's and therefore do not show up on the PR curve. To elaborate, a DTBB is said to match a GTBB if 
\begin{enumerate}
    
\item its IoU with this GTBB is greater than the IoU threshold $t$, and

\item its IoU with this GTBB is greater than its IoU with any other GTBB from the same class.
    
\end{enumerate}
A DTBB is an FP if it does not have an IoU greater than $t$ with any GTBB from the same class. Some benchmarks, such as Microsoft COCO, use a list of IoU thresholds and take the average of the mAP's. Here we follow Pascal VOC and use a single threshold to demonstrate the process of computing mAP. The same process can be extended to handle a list of IoU thresholds.

A DTBB is also counted as an FP if the GTBB it matches has already been matched by another DTBB that has a higher confidence score; a DTBB is ignored if it matches a "difficult" GTBB no matter if this GTBB has been matched or not; a DTBB is a TP if it is neither an FP nor ignored.
\section{Related Work}\label{section_related_work}
The original MATLAB script to compute mAP is released along with the Pascal VOC development kit. \cite{faster_rcnn} later provides a Python version in their Faster R-CNN repository. Both implementations require dumping every DTBB (after post-processing) to a file depending on each of their class label. A significant amount of time is thus spent on transferring detection results from memory to disk. \cite{detectron2} further improve the Python implementation by getting rid of the I/O between memory and disk. However, the rest of their code remains the same as the original implementation. 

These sequential implementations cannot directly process outputs from neural networks in parallel. Therefore, they are not suitable for quickly showing the progress on mAP after each training epoch. They are designed to compute mAP in two phases after inference is made for an entire dataset.

\subsection{Collecting Precision and Recall
Data}\label{subsection_collecting_precision_and_recall_data}
The objective of the first phase is to collect precision and recall data for each class. After parsing the annotation files and detection results, we have
\begin{itemize}

\item $\sB$, a list that contains $k$ matrices, $\{ \mB_0 , \mB_1, \dotsc, \mB_{k-1} \} \in \R^{\{ z_0 , z_1, \dotsc, z_{k - 1} \} \times 4}$, where $z_i$ is the total number of DTBB's from the $i$-th class. Each row of a matrix is a vector of length 4 representing the four coordinates, $x^\text{min}$, $y^\text{min}$, $x^\text{max}$, and $y^\text{max}$, of a DTBB.

\item $\sJ$, a list that contains $k$ vectors, $\{ \vj_0 , \vj_1 , \dotsc, \vj_{k-1} \} \in \R^{ \{ z_0, z_1, \dotsc, z_{k-1} \}}$. Each element of a vector is an integer representing the index of the image that its corresponding DTBB belongs to. The value of an image index ranges from $0$ to $g - 1$, where $g$ is the total number of images in the dataset.

\item $\sS$, a list that contains $k$ vectors, $\{ \vs_0 , \vs_1 , \dotsc, \vs_{k-1} \} \in \R^{ \{ z_0, z_1, \dotsc, z_{k-1} \}}$. Each element of a vector is a floating point number representing the confidence score of its corresponding DTBB. The value of a confidence score ranges from $0$ to $1$.

\item $\hat{\sB}$, a list that contains $g$ matrices, $\{ \hat{\mB}_0 , \hat{\mB}_1, \dotsc, \hat{\mB}_{g-1} \} \in \R^{\{ \hat{m}_0 , \hat{m}_1, \dotsc, \hat{m}_{g-1} \} \times 4}$, where $\hat{m}_i$ is the number of GTBB's in the $i$-th image. Each row of a matrix is a vector of length 4 representing the four coordinates, $x^\text{min}$, $y^\text{min}$, $x^\text{max}$, and $y^\text{max}$, of a GTBB.

\item $\hat{\sL}$, a list that contains $g$ vectors, $\{ \hat{\vl}_0 , \hat{\vl}_1, \dotsc, \hat{\vl}_{g-1} \} \in \R^{ \{ \hat{m}_0, \hat{m}_1, \dotsc, \hat{m}_{g-1} \}}$. Each element of a vector is an integer representing the index of the class its corresponding GTBB belongs to.

\item $\hat{\sF}$, a list that contains $g$ vectors, $\{ \hat{\vf}_0 , \hat{\vf}_1, \dotsc, \hat{\vf}_{g-1} \} \in \R^{ \{ \hat{m}_0, \hat{m}_1, \dotsc, \hat{m}_{g-1} \}}$. Each element of a vector is a boolean value indicating if its corresponding GTBB is "difficult". A value of $1$ means the corresponding GTBB is "difficult", and $0$ means otherwise.

\item $\hat{\sD}$, a list that contains $g$ vectors, $\{ \hat{\vd}_0 , \hat{\vd}_1, \dotsc, \hat{\vd}_{g-1} \} \in \R^{ \{ \hat{m}_0, \hat{m}_1, \dotsc, \hat{m}_{g-1} \}}$. Each element of a vector is a boolean value indicating if its corresponding GTBB has been matched. A value of $1$ means the corresponding GTBB has been matched, and $0$ means otherwise.

\item $\hat{\vz} \in \R^k$, a vector that contains $k$ integers. An element $\hat{z}[ i ]$ represents the number of "easy" GTBB's from the $i$-th class.

\end{itemize}
Sequential implementations use a nested loop to compute mAP. Each time the outer loop is executed, all DTBB's from a specific class are sorted by their confidence scores and are then passed to the inner loop. Two temporary vectors are also created each time, one is to keep track of TP's, and the other is to keep track of FP's.

In the inner loop, one DTBB is processed at a time.  First, all GTBB's that are from the same image and the same class as the DTBB are extracted. Second, the IoU's between the DTBB and the extracted GTBB's are calculated. Finally, the DTBB is categorized as ignored, TP, or FP based on the rules mentioned in Sec. \ref{section_review}.

After all DTBB's from the class under investigation are categorized, the inner loop is exited. Precision and recall values are calculated in the outer loop using the cumulative sums of TP's and FP's.

Alg. \ref{algorithm_sequential_map_computation_collect_precision_and_recall_data} shows the pseudocode of collecting precision and recall data.
\begin{algorithm}[tb]

\caption{Sequential mAP Computation - Collecting Precision and Recall Data}

\begin{algorithmic}\label{algorithm_sequential_map_computation_collect_precision_and_recall_data}

\STATE {\bfseries Input:} $\sB, \sJ, \sS, \hat{\sB}, \hat{\sL}, \hat{\sF}, \hat{\sD}, \hat{\vz}, t, \epsilon$ \# $\epsilon$ is used to prevent division by zero

\STATE $\sP = \{ \}$ \# precision

\STATE $\sQ = \{ \}$ \# recall

\FOR{$i \ \textbf{in} \ \texttt{range} ( \texttt{len}  ( \sS ) )$}

 \STATE Sort $\sB [ i ]$ and $\sJ [ i ]$ by $\sS [ i ]$ \# sort by confidence scores

 \STATE $\mB = \sB [ i ]$
 
 \STATE $\vs = \sS [ i ]$
 
 \STATE $\vu = \texttt{zeros\_like}( \vs )$ \# TP's
 
 \STATE $\vv = \texttt{zeros\_like}( \vs )$ \# FP's
 
 \FOR{$j \ \textbf{in} \ \texttt{range} ( \texttt{len} ( \vs ) )$}
 
  \STATE $\vb = \mB [ j ]$
 
  \STATE $\hat{i} = \sJ [ i ] [ j ]$ \# the image index of the $j$-th DTBB from the $i$-th class
  
  \STATE $\hat{\vj} = \hat{\sL} [ \hat{i} ] == i$ \# indices of GTBB's from the $i$-th class in the $\hat{i}$-th image 
 
  \STATE $\hat{\mB} = \hat{\sB} [ \hat{i} ] [ \hat{\vj} ]$
  
  \STATE $\text{IoU} = -\infty$
  
  \IF{$\texttt{len} ( \hat{\mB} ) > 0$}
   
   \STATE $\vw = \texttt{maximum} ( 0 , \texttt{minimum} ( \vb [ 2 ] , \hat{\mB} [ : , 2 ] ) - \texttt{maximum} ( \vb [ 0 ] , \hat{\mB} [ : , 0 ] ) + 1 )$
   
   \STATE $\vh = \texttt{maximum} ( 0 , \texttt{minimum} ( \vb [ 3 ] , \hat{\mB} [ : , 3 ] ) - \texttt{maximum} ( \vb [ 1 ] , \hat{\mB} [ : , 1 ] ) + 1 )$
   
   \STATE $a =  ( \vb [ 2 ] - \vb [ 0 ] + 1 ) \cdot ( \vb [ 3 ] - \vb [ 1 ] + 1 ) )$
   
   \STATE $\hat{\va} =  ( \hat{\mB} [ : , 2 ] - \hat{\mB} [ : , 0 ] + 1 ) \cdot ( \hat{\mB} [ : , 3 ] - \hat{\mB} [ : , 1 ] + 1 )$
   
   \STATE $\vo = ( \vw \cdot \vh )  / ( a + \va - \vw \cdot \vh  )$
   
   \STATE $\text{IoU} = \texttt{amax} ( \vo )$
   
   \STATE $\hat{k} = \texttt{argmax} ( \vo )$
   
  \ENDIF
  
  \IF{$\text{IoU} > t$}
  
    \IF{$\hat{\sF} [ \hat{i} ] [ \hat{\vj} ] [ \hat{k} ] == 0$}
    
     \IF{$\hat{\sD} [ \hat{i} ] [ \hat{\vj} ] [ \hat{k} ] == 0$}
     
      \STATE $\vu [ j ] = 1$
      
      \STATE $\hat{\sD} [ \hat{i} ] [ \hat{\vj} ] [ \hat{k} ] = 1$
      
     \ELSE
     
      \STATE $\vv [ j ] = 1$
      
     \ENDIF
     
    \ENDIF
    
   \ELSE
   
    \STATE $\vv [ j ] = 1$
    
   \ENDIF
   
 \ENDFOR
 
 \STATE $\vp = \texttt{cumsum} ( \vu ) / \hat{\vz} [ i ]$
 
 \STATE $\vq = \texttt{cumsum} ( \vu ) / \texttt{maximum} ( \texttt{cumsum} ( \vu ) + \texttt{cumsum} ( \vv ), \epsilon)$
 
 \STATE Add $\vp$ to $\sP$
 
 \STATE Add $\vq$ to $\sQ$
 
\ENDFOR

\STATE {\bfseries Output:} $\sP, \sQ$

\end{algorithmic}

\end{algorithm}
We observe the following problems:
\begin{enumerate}

\item The outer loop shows that DTBB's from different classes are not processed jointly. The reason is that we only want to match DTBB's and GTBB's from the same class. A human DTBB with 100\% confidence, for instance, cannot be a TP even if it perfectly matches a dog GTBB. 

\item The inner loop shows that for each image, only one DTBB is processed at a time. The reason is that judging if a DTBB is a TP depends on if the target GTBB has already been matched by some previous DTBB whose confidence score is higher. 

\end{enumerate}
\subsection{Calculating mAP}\label{subsection_calculating_map}
Precision and recall data are collected for every class in the previous phase. The current phase's goal is to calculate AP for each class and take the mean to obtain the mAP. One can either use every point on the PR curve to calculate AP or use specific points that correspond to a sequence of equally spaced recall levels ranging from 0 to 1.

\subsubsection{Using Every Point}\label{subsubsection_using_every_point}
If the former method is used, spikes on the PR curve are flattened by changing precision values in reverse order according to Equ. \ref{equation_calculating_map_method_1},
\begin{equation}\label{equation_calculating_map_method_1}
p_{i-1} = p_i \text{ if } p_{i - 1} <  p_i
\end{equation}
where $i \in \{ 1, 2, \dotsc, z - 1 \}$, and $z$ is the total number of DTBB's from the class under investigation.

\begin{figure}[tb]

\vskip 0.2in

\begin{center}

\centerline{\includegraphics[width=\columnwidth]{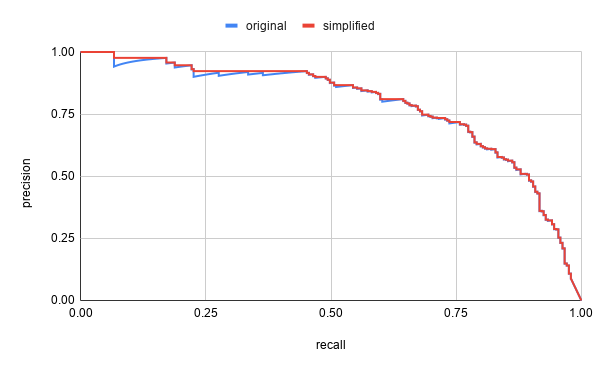}}

\caption{Original PR curve vs. simplified PR curve. The blue line shows the original curve, and the red line shows the simplified curve. The area under the simplified curve is equivalent to the sum of the rectangular areas that correspond to the steps.}

\label{figure_zig_zag}

\end{center}

\vskip -0.2in

\end{figure}

After changing the precision values, the original PR curve is simplified to a step curve, as shown in Fig. \ref{figure_zig_zag}.
AP is then calculated by summing the rectangular areas that correspond to the steps. Alg. \ref{algorithm_calculating_map_using_every_point} shows the pseudocode of calculating mAP using every point on PR curves.
\begin{algorithm}[tb]

\caption{Sequential mAP Computation - Calculating mAP Using Every Point}

\begin{algorithmic}\label{algorithm_calculating_map_using_every_point}

\STATE {\bfseries Input:} $\sP$, $\sQ$

\STATE $\text{mAP} = 0$

\FOR{$i \ \textbf{in} \ \texttt{range} ( \texttt{len} ( \sP ) )$}
 
 \STATE $\vp = \texttt{concatenate} ( ( [ 0.0 ] , \sP [ i ] , [ 0.0 ] ) )$
 
 \STATE $\vq = \texttt{concatenate} ( ( [ 0.0 ] , \sQ [ i ] , [ 1.0 ] ) )$
 
 \STATE $\text{AP} = 0$
 
 \FOR{$j \ \textbf{in} \ \texttt{range}( \texttt{len} ( \vp ) - 1, 1, -1 )$}
  
  \STATE $\vp [ j - 1 ] = \texttt{maximum} ( \vp [ j - 1 ] , \vp [ j ] )$ 
 
 \ENDFOR
 
 \STATE $\vj = \texttt{where} ( \vq [ 1 : ] \neq \vq [ : -1 ] )$ \# a vector of indices where recall values change
 
 \STATE $\text{AP} = \texttt{sum} ( \vp [ \vj + 1 ] \cdot ( \vq [ \vj + 1 ] - \vq [ \vj ] ) )$
 
 \STATE $\text{mAP} = \text{mAP} + \text{AP} / \texttt{len} ( \sP )$

\ENDFOR

\STATE {\bfseries Output:} mAP

\end{algorithmic}

\end{algorithm}
\subsubsection{Using Specific Points}\label{subsubsection_using_specific_points}
If the latter method is used, mAP is calculated by taking the arithmetic average of the precision values at specific recall levels (e.g., $\{ 0.0, 0.1, \dotsc 1.0 \}$). The precision value $p_1$ at a specific recall level $q_1$ is defined as the maximum precision value $p_2$ at some recall level $q_2$ satisfying $q_2 \geq q_1$. Alg. \ref{algorithm3} shwos the pseudocode of calculating mAP using specific points on PR curves.
\begin{algorithm}[tb]

\caption{Sequential mAP Computation - Calculating mAP Using Specific Points}

\begin{algorithmic}\label{algorithm3}

\STATE {\bfseries Input:} $\sP$, $\sQ$, $\hat{\vq}$

\STATE $\text{mAP} = 0$

\FOR{$i \ \textbf{in} \ \texttt{range} ( \texttt{len} ( \sP ) )$}
 
 \STATE AP $= 0$
 
 \FOR{$\hat{q} \ \textbf{in} \ \hat{\vq}$}
  
  \STATE $\vj = \sQ [ i ] \geq \hat{q}$ \# a binary mask indicating if elements of $\sQ [ i ]$ are greater than $\hat{q}$
  
  \IF{$\texttt{sum} ( \vj ) > 0$}
  
  \STATE $\text{AP} = \text{AP} + \texttt{amax} ( \sP [ i ] [ \vj ] ) / \texttt{len} ( \hat{\vq} )$
  
  \ENDIF
 
 \ENDFOR
 
 \STATE $\text{mAP} = \text{mAP} + \text{AP} / \texttt{len} ( \sP )$

\ENDFOR

\STATE {\bfseries Output:} mAP

\end{algorithmic}

\end{algorithm}
\section{Method}\label{section_method}
One major problem we see from sequential implementations is that matrices and vectors involved in computing mAP generally do not have fixed dimensions. For example, the numbers of GTBB's per image are often not the same in a batch of images. Even in the same image, the numbers of GTBB's per class are not likely to be the same.

However, current machine learning software and hardware accelerators expect regular input data whose lengths are fixed along each axis. Earlier implementations fail to parallelize the computation of mAP because they see it as an isolated process after inference. We argue that an essential step to accelerate the computation of mAP is to revisit the steps before it and locate where irregular input dimensions occur.
\subsection{Data Loading}\label{subsection_data_loading}
Sequential implementations yield one image at a time in the data loading stage. The corresponding annotation file is not parsed until all images have gone through the detector. On the other hand, our implementation processes a mini-batch of images and their corresponding annotation files on the fly.

For a mini-batch of $n$ samples, we have
\begin{itemize}

\item $\sX$, a list that contains $n$ tensors, $\{ \tX_0, \tX_1, \dotsc, \tX_{n-1} \}$. A tensor $\tX_i \in \R^{c_i \times h_i \times w_i}$ represents the pixel values of the $i$-th image, where $c_i$ is the number of channels, $h_i$ is the height, and $w_i$ is the width.

\item $\hat{\sB}$, a list that contains $n$ matrices, $\{ \hat{\mB}_0, \hat{\mB}_1, \dotsc, \hat{\mB}_{n-1} \}$. A matrix $\hat{\mB}_i \in \R^{\hat{m}_i \times 4}$ represents all GTBB's in the $i$-th image, where $\hat{m}_i$ is the number of GTBB's in the $i$-th image.

\item $\hat{\sL}$, a list that contains $n$ vectors, $\{ \hat{\vl}_0, \hat{\vl}_1, \dotsc, \hat{\vl}_{n - 1} \}$. A vector $\hat{\vl}_i \in \R^{\hat{m}_i}$ represents the class labels of all GTBB's in the $i$-th image.

\item $\hat{\sM}$, a list that contains $n$ vectors, $\{ \hat{\vm}_0, \hat{\vm}_0, \dotsc, \hat{\vm}_{n-1} \}$. A vector $\hat{\vm}_i$ represents if the corresponding GTBB's in the $i$-th image are "difficult".
    
\end{itemize}
Our implementation converts each of the above lists to an array that has fixed length along each axis. For $\sX$, the typical approach is to pre-process each image such that all of them have the same size. Nevertheless, the pre-processing can be different for each object detection algorithm; therefore, we do not go into further details about pre-processing $\sX$.

As for $\hat{\sB}$, $\hat{\sL}$, and $\hat{\sM}$, we need to solve the problem that for any $i, j \in \{ 0, 1, \dotsc, n - 1 \}$ and $i \neq j$, we have $\hat{m}_i \neq \hat{m}_j$ in general. One way to enforce an equal number of GTBB's is to add dummy GTBB's to each image such that all of them have $\hat{m}$ GTBB's, where $\hat{m}$ is the maximum number of GTBB's per image in the mini-batch (or dataset).

To keep track of dummy GTBB's, we create a matrix of boolean values, $\hat{\mM} \in \R^{n \times \hat{m}}$, whose elements indicate if the corresponding GTBB's are real. We can further overwrite the elements of $\hat{\mM}$ with the corresponding elements of $\hat{\sM}$ such that a value of $1$ means DTBB's matching the corresponding GTBB will be ignored, and $0$ means otherwise.

Alg. \ref{algorithm_data_loading} shows the pseudocode of our data loading stage (without the part that pre-processes input images).
\begin{algorithm}[tb]

\caption{Parallel mAP Computation - Data Loading}

\begin{algorithmic}\label{algorithm_data_loading}

\STATE {\bfseries Input:} $\hat{\sB}, \hat{\sL}, \hat{\sM}, n, \hat{m}$

\STATE $\hat{\tB} = \texttt{zeros} ( ( n , \hat{m} , 4 ) )$

\STATE $\hat{\mL} = \texttt{zeros} ( ( n , \hat{m} ) )$

\STATE $\hat{\mM} = \texttt{ones} ( ( n , \hat{m} ) )$

\FOR{$i \ \textbf{in} \ \texttt{range} ( n )$}

 \STATE $j = \texttt{len} ( \hat{\sB} [ i ] )$
 
 \STATE $\hat{\tB} [ i , : j ] = \hat{\sB} [ i ]$
 
 \STATE $\hat{\mL} [ i , : j ] = \hat{\sL} [ i ]$

 \STATE $\hat{\mM} [ i , : j ] = \hat{\sM} [ i ]$ 
 
\ENDFOR

\STATE {\bfseries Output:} $\hat{\tB}, \hat{\mL}, \hat{\mM}$

\end{algorithmic}

\end{algorithm}
For each mini-batch, our data loader outputs
\begin{itemize}

\item $\hat{\tB} \in \R^{n \times \hat{m} \times 4}$, a rank 3 tensor representing all GTBB's in the mini-batch.

\item $\hat{\mL} \in \R^{n \times \hat{m}}$, a matrix representing the class labels of all GTBB's in the mini-batch.

\item $\hat{\mM} \in \R^{n \times \hat{m}}$, a matrix whose elements indicate if DTBB's matching the corresponding GTBB's will be ignored.
    
\end{itemize}
\subsection{Post-Processing}\label{subsection_post_processing}
Modern object detectors that are convolutional neural networks (CNN's) usually produce a fixed number of DTBB's per image. However, many of them, such as SSD \citep{ssd} and RetinaNet \citep{retinanet}, apply non-maximum suppression (NMS) \citep{nms} or other post-processing techniques to eliminate unwanted DTBB's. By doing so, the numbers of DTBB's are likely to be different for images in a batch. To prevent this from happening, our implementation keeps the discarded DTBB's and uses a matrix of boolean values to keep track of them.

Just before post-processing, we have 
\begin{itemize}

\item $\tB \in \R^{n \times m \times 4}$, a rank 3 tensor representing all DTBB's in the mini-batch, where $n$ is the batch size, and $m$ is the number of DTBB's per image.

\item $\mL \in \R^{n \times m}$, a matrix representing the labels of all DTBB's in the mini-batch.
    
\item $\mS \in \R^{n \times m}$, a matrix representing the confidence scores of all DTBB's in the mini-batch.
    
\end{itemize}
We create an additional matrix $\mM \in \R^{n \times m}$. If an element $\mM [ i , j ]$ is equal to 1, it means that the $j$-th DTBB in the $i$-th image is discarded. If instead $\mM [ i , j ]$ is equal to 0, it means the corresponding DTBB is kept after post-processing. Alg. \ref{algorithm_post_processing} shows our post-processing stage.
\begin{algorithm}[tb]

\caption{Parallel mAP Computation - Post-Processing}

\begin{algorithmic}\label{algorithm_post_processing}

\STATE {\bfseries Input:} $\tB, \mL, \mS, n, m$

\STATE $\mM = \texttt{zeros} ( ( n , m ) )$

\STATE Post-process $\tB$, $\mL$, $\mS$

\FOR{$i \ \textbf{in} \ \texttt{range} ( n )$}

 \FOR{$j \ \textbf{in} \ \texttt{range} ( m )$}
 
  \IF{$\tB [ i , j ], \mL [ i , j ], \mS [ i , j ]$ are supposed to be discarded}
  
   \STATE $\mM [ i , j ] = 1$
   
  \ENDIF
  
 \ENDFOR
 
\ENDFOR

\STATE {\bfseries Output:} $\tB, \mL, \mS, \mM$

\end{algorithmic}

\end{algorithm}
\subsection{IoU Computation}\label{subsection_iou_computation}
Sequential implementations only calculate the IoU's between one DTBB and all GTBB's that are from the same image and the same class. Our implementation takes one step further and calculates the IoU's between every DTBB and every GTBB in the same image. It also does this simultaneously for a mini-batch of images.

Given a mini-batch of DTBB's, $\tB \in \R^{n \times m \times 4}$, and a mini-batch of GTBB's, $\hat{\tB} \in \R^{n \times \hat{m} \times 4}$, we have $m \neq \hat{m}$ in general. However, parallel implementations of computing IoU largely depend on element-wise operations, which require the two operands have the same shape. One solution is to expand each tensor along its missing axis.

For example, $\tB$ can be expanded to $\underline{\tB} \in \R^{n \times m \times \hat{m} \times 4}$ by repeating its elements $\hat{m}$ times along the inserted $2$-nd axis. Similarly, $\hat{\tB}$ can be expanded to $\underline{\hat{\tB}} \in \R^{n \times m \times \hat{m} \times 4}$ by repeating its elements $m$ times along the inserted $1$-st axis. Alg. \ref{algorithm_iou_computation} shows the pseudocode of our IoU computation stage.
\begin{algorithm}[tb]

\caption{Parallel mAP Computation - IoU Computation}

\begin{algorithmic}\label{algorithm_iou_computation}

\STATE {\bfseries Input:} $\tB, \hat{\tB}, n, m, \hat{m}$

\STATE $\underline{\tB} = \texttt{tile} ( \texttt{expand\_dims} ( \tB , \texttt{axis} \texttt{=} 2 ), ( 1 , 1 , \hat{m} , 1 ) )$

\STATE $\underline{\hat{\tB}} = \texttt{tile} ( \texttt{expand\_dims} ( \hat{\tB} , \texttt{axis} \texttt{=} 1 ), ( 1 , m , 1 , 1 ) )$

\STATE $\tW = \texttt{maximum} ( 0 , \texttt{minimum} ( \underline{\tB} [ \dots , 2 ] , \underline{\hat{\tB}} [ \dots , 2 ] ) - \texttt{maximum} ( \underline{\tB} [ \dots , 0 ] , \underline{\hat{\tB}} [ \dots , 0 ] ) + 1 )$

\STATE $\tH = \texttt{maximum} ( 0 , \texttt{minimum} ( \underline{\tB} [ \dots , 3 ] , \underline{\hat{\tB}} [ \dots , 3 ] ) - \texttt{maximum} ( \underline{\tB} [ \dots , 1 ] , \underline{\hat{\tB}} [ \dots , 1 ] ) + 1 )$

\STATE $\tA = ( \underline{\tB} [ \dots , 2 ] - \underline{\tB} [ \dots , 0 ] + 1 ) \cdot ( \underline{\tB} [ \dots , 3 ] - \underline{\tB} [ \dots , 1 ] + 1 )$

\STATE $\hat{\tA} = ( \underline{\hat{\tB}} [ \dots , 2 ] - \underline{\hat{\tB}} [ \dots , 0 ] + 1 ) \cdot ( \underline{\hat{\tB}} [ \dots , 3 ] - \underline{\hat{\tB}} [ \dots , 1 ] + 1 )$

\STATE $\tO = ( \tW \cdot \tH ) / ( \tA + \hat{\tA} - \tW \cdot \tH ) $

\STATE {\bfseries Output:} $\tO$

\end{algorithmic}

\end{algorithm}
\subsection{IoU Filtering}\label{subsection_iou_filtering}
Although our implementation can compute $\tO \in \R^{n \times m \times \hat{m}}$, the IoU's between every DTBB and every GTBB for a mini-batch of images, it may "mistakenly" compute some invalid IoU's. An IoU is invalid if
\begin{enumerate}

\item the DTBB and the GTBB have different class labels, or

\item the DTBB is discarded after post-processing.

\end{enumerate}
We set any invalid IoU's to 0 so that they do not affect determining TP's. However, doing so will erroneously mark discarded DTBB's as FP's when they should be ignored. This "mistake" is made intentionally and will be corrected in Sec. \ref{subsubsection_below_threshold_dtbbs}. Alg. \ref{algorithm_iou_filtering} shows the pseudocode of our IoU filtering stage.
\begin{algorithm}[tb]

\caption{Parallel mAP Computation - IoU Filtering}

\begin{algorithmic}\label{algorithm_iou_filtering}

\STATE {\bfseries Input:} $\tO, \mL, \hat{\mL}, \mM, m, \hat{m}$

\STATE $\tL = \texttt{tile} ( \texttt{expand\_dims} ( \mL , \texttt{axis} \texttt{=} 2 ), ( 1 , 1 , \hat{m} ) )$

\STATE $\hat{\tL} = \texttt{tile} ( \texttt{expand\_dims} ( \hat{\mL} , \texttt{axis} \texttt{=} 1 ), ( 1 , m , 1 ) )$

\STATE $\tO [ \tL \neq \hat{\tL} ] = 0$

\STATE $\tM = \texttt{tile} ( \texttt{expand\_dims} ( \mM , \texttt{axis} \texttt{=} 2 ), ( 1 , 1 , \hat{m} ) )$

\STATE $\tO [ \tM ] = 0$

\STATE {\bfseries Output:} $\tO$

\end{algorithmic}

\end{algorithm}
\subsection{DTBB categorization}\label{subsection_dtbb_categorization}
Given a mini-batch of $n \times m$ DTBB's, we create a matrix $\mD \in \R^{n \times m}$ to categorize each DTBB. An element of $\mD$ can have a value of
\begin{itemize}
    
\item $0$ if the corresponding DTBB is ignored, or
    
\item $1$ if the corresponding DTBB is an FP, or
    
\item $2$ if the corresponding DTBB is a TP.
    
\end{itemize}
Our implementation categorizes DTBB's in two steps. In the first step, we focus on DTBB's that do not meet the IoU threshold requirement. In the second step, we focus on the rest of DTBB's.
\subsubsection{Below-Threshold DTBB's}\label{subsubsection_below_threshold_dtbbs}
For DTBB's that do not meet the IoU threshold requirement, we first set their corresponding elements in $\mD$ to 1.
Doing so will lead to the "mistake" mentioned in Sec. \ref{subsection_iou_filtering}. We then correct this "mistake" by setting elements of $\mD$ to 0, where the corresponding elements of $\mM$ are equal to 1. Alg. \ref{algorithm_below_threshold_dtbb_categorization} shows the pseudocode of our categorization stage for below-threshold DTBB's.
\begin{algorithm}[tb]

\caption{Parallel mAP Computation - Below-Threshold DTBB Categorization}

\begin{algorithmic}\label{algorithm_below_threshold_dtbb_categorization}

\STATE {\bfseries Input:} $\tO, \mM, n, m, t$

\STATE $\mD = \texttt{zeros} ( n , m )$

\STATE $\mD [ \texttt{amax} ( \tO , \texttt{axis} \texttt{=} 2 ) \leq t ] = 1$

\STATE $\mD [ \mM ] = 0$

\STATE {\bfseries Output:} $\mD$

\end{algorithmic}

\end{algorithm}
\subsubsection{Above-Threshold DTBB's}
DTBB's that meet the IoU threshold requirement are processed differently based on whether or not they match "difficult" GTBB's. A DTBB is ignored during evaluation if it matches a "difficult" GTBB. For such DTBB's, we set their corresponding elements in $\tO$ to 0.

Now the only DTBB's left are the ones whose corresponding elements in $\tO$ are greater than $t$. Each of these DTBB's can either be
\begin{itemize}

\item a TP, if it matches an "easy" GTBB, and has the highest confidence score among all DTBB's matching the same GTBB, or

\item an FP, if it matches an "easy" GTBB, but does not have the highest confidence score among all DTBB's matching the same GTBB.

\end{itemize}
Our implementation first sets elements of $\mD$ to 1, where the corresponding elements in $\tO$ are greater than $t$. Doing so "mistakenly" marks all above-threshold DTBB's that match "easy" GTBB's as FP's. We then correct this "mistake" by finding the TP's and setting their corresponding elements in $\mD$ to 2.

To elaborate, we first obtain $\mV \in \R^{n \times m}$, the FP mask for above-threshold DTBB's, by comparing the maximum elements of $\tO$ along the $2$-nd axis against $t$. At the same time, we also get $\mJ \in \R^{n \times m}$, a matrix that contains the indices where maximum elements occur. We then set elements of $\mD$ to 1, where the corresponding elements in $\mV$ are equal to 1. The only thing left to do now is finding the TP's.

Because a DTBB must have the highest confidence score among all DTBB's matching the same "easy" GTBB for it to be a TP, we sort $\mD$, $\mV$, and $\mJ$ by $\mS$. Some post-processing techniques may sort DTBB's by their confidence scores. If such post-processing techniques have been adopted, this step can be skipped.

Given sorted $\mJ$, sequential implementations find TP's by iterating over its elements along the $1$-st axis. Hash tables are often used to keep track of GTBB's that have been matched. In each iteration, sequential implementations search the correct hash table to see if there is a key equal to the current element of $\mJ$. If there is, then the corresponding DTBB is marked as an FP; otherwise, the corresponding DTBB is marked as a TP.

We observe, for any TP whose corresponding element $\mJ [ i , j_1 ]$ is equal to $\hat{j}$, the condition in Equ. \ref{equation_sorted_j_property} must be true.
\begin{equation}\label{equation_sorted_j_property}
j_1 \leq j_2 \ \forall \ j_2 \in \{ 0, 1, \dotsc, m - 1 \} \ s.t. \ J [ i , j_1 ] = J [ i, j_2 ] = \hat{j}
\end{equation}
In other words, if the $j_1$-th DTBB is a TP, $j_1$ must be smaller than or equal to any $j_2$ such that the $j_1$-th DTBB and the $j_2$-th DTBB have the same class label $\hat{j}$. Of course, both DTBB's must be from the same $i$-th image of a mini-batch. This property of sorted $\mJ$ is the foundation for breaking the data dependency in sequential implementations. Our implementation takes advantage of this property and transforms $\mJ$ into a tensor whose minimum values along the $2$-nd axis correspond to TP's.

To begin with, we transform $\mJ$ to its one-hot encoding $\tJ \in \R^{n \times m \times \hat{m}}$. Next, we expand $\mV$ to $\tV \in \R^{n \times m \times \hat{m}}$ so that $\tV$ and $\tJ$ have the same shape. We then set elements of $\tJ$ to 0, where the corresponding elements in $\tV$ are equal to 0. Doing so filters out DTBB's that have already been categorized in previous steps. A $0$ in $\tJ$ means the corresponding DTBB has already been categorized, and a $1$ means otherwise.  

For convenience, we swap the $1$-st and $2$-nd axes of $\tJ$ so that its shape is $(n , \hat{m}, m)$. Finding TP's is now equivalent to finding the first 1's in the $2$-nd axis of $\tJ$. Our implementation treats finding the first 1's as finding the minimum values to take advantage of optimized operators.

We multiply $\tJ$ with $\vn \in \R^m$, a vector that contains a range of numbers, $\{ 1, 2, \dotsc, m \}$. Each 2-D slice $\tJ [ : , : , j ]$ is multiplied with $j + 1$ so that 0's in $\tJ$ remain 0's and 1's in $\tJ$ are set to their indices plus one. Since we are interested in finding the minimum values, we set all 0's in $\tJ$ to some sentinel value $\lambda$, satisfying $\lambda > m$. Now an element $\tJ [ i , \hat{j} , j ]$ corresponds to a TP if
\begin{enumerate}

\item it is equal to the minimum value in the 1-D slice $\tJ [ i , j , : ]$, and

\item it is smaller than $\lambda$

\end{enumerate}
We get a tensor of boolean values, $\tU \in \R^{n \times \hat{m} \times m}$, by checking if each element of $\tJ$ satisfies both the above conditions. The TP mask, $\mU \in \R^{n \times m}$, can then be obtained by finding the maximum values of $\tU$ along the $1$-st axis. Finally, we set elements of $\mD$ to 2, where the corresponding elements in $\mU$ are equal to 1. Alg. \ref{algorithm_above_threshold_dtbb_categorization} shows the pseudocode of our categorization stage for above-threshold DTBB's.
\begin{algorithm}[tb]

\caption{Parallel mAP Computation - Above-Threshold DTBB Categorization}

\begin{algorithmic}\label{algorithm_above_threshold_dtbb_categorization}

\STATE {\bfseries Input:} $\mD, \tO, \mS, \hat{\mM}, m, \hat{m}, t, \lambda$

\STATE $\hat{\tM} = \texttt{tile} ( \texttt{expand\_dims} ( \hat{\mM} , \texttt{axis} \texttt{=} 1 ), ( 1 , m , 1 ) )$

\STATE $\tO [ \hat{\tM} ] = 0$

\STATE $\mV = \texttt{amax} ( \tO , \texttt{axis} \texttt{=} 2 ) > t$ \# FP mask

\STATE $\mJ = \texttt{argmax} ( \tO , \texttt{axis} \texttt{=} 2 )$

\STATE Sort $\mD, \mV, \mJ$ by $\mS$ (if have not done)

\STATE $\mD [ \mV ] = 1$

\STATE $\tJ = \texttt{tile} ( \texttt{expand\_dims} ( \mJ , \texttt{axis} \texttt{=} 2 ), ( 1 , 1 , \hat{m} ) ) == \texttt{arange} ( \hat{m} )$ \# one-hot encode $\mJ$

\STATE $\tV = \texttt{tile} ( \texttt{expand\_dims} ( \mV , \texttt{axis} \texttt{=} 2 ), ( 1 , 1 , \hat{m} ) )$

\STATE $\tJ [ \sim \tV  ] = 0$

\STATE $\tJ = \texttt{transpose} ( \tJ , \texttt{axes} \texttt{=} ( 0 , 2 , 1 ) )$

\STATE $\tJ =  \tJ \cdot ( \texttt{arange} ( m ) + 1 )$ \# not $\texttt{arange} ( m  + 1)$

\STATE $\tJ [ \tJ == 0 ] = \lambda$

\STATE $\tU = (\tJ == \texttt{amin} ( \tJ , \texttt{keepdims} \texttt{=} \texttt{True} ) ) \ \& \ ( \tJ < \lambda )$

\STATE $\mU = \texttt{amax} ( \tU , \texttt{axis} \texttt{=} 1 )$ \# TP mask

\STATE $\mD [ \mU ] = 2$

\STATE {\bfseries Output:} $\mD$

\end{algorithmic}

\end{algorithm}
\subsection{TP and FP Extraction}\label{subsection_tp_and_fp_extraction}
After setting each D's element to the correct value, we find indices where the corresponding elements are greater than 0. Each of these indices corresponds to either a TP or an FP. We use these indices to extract the corresponding elements in $\mL$ and $\mS$. We also extract the corresponding elements in $\mD$. However, we subtract the values of the extracted elements by 1 so that they form a binary mask. A value of 1 means the subtracted element corresponds to a TP, and 0 means otherwise. Extracted elements from $\mL$, $\mS$, and $\mD$ are added to their corresponding lists, $\sL$, $\sS$, and $\sD$. Alg. \ref{algorithm_tp_and_fp_extraction} shows the pseudocode of our TP and FP extraction stage.
\begin{algorithm}[tb]

\caption{Parallel mAP Computation - TP and FP Extraction}

\begin{algorithmic}\label{algorithm_tp_and_fp_extraction}

\STATE {\bfseries Input:} $\sL, \sS, \sD$

\STATE Add $\mL[ \mD > 0 ]$ to $\sL$

\STATE Add $\mS[ \mD > 0 ]$ to $\sS$

\STATE Add $\mD[ \mD > 0 ] - 1$ to $\sD$

\STATE {\bfseries Output:} $\sL, \sS, \sD$

\end{algorithmic}

\end{algorithm}
\subsection{PR Computation}\label{subsection_pr_computation}
After all mini-batches of samples have been processed, we have
\begin{itemize}

\item $\sL$, a list that contains $\texttt{ceil} ( g / n )$ vectors, where $g$ is the total number of images in the dataset, and $n$ is the batch size. Each vector contains the class labels of DTBB's that are not ignored in its corresponding mini-batch.

\item $\sS$, a list that contains $\texttt{ceil} ( g / n )$ vectors. Each vector contains the confidence scores of DTBB's that are not ignored in its corresponding mini-batch.

\item $\sD$, a list that contains $\texttt{ceil} ( g / n )$ vectors. Each vector contains the boolean values for DTBB's that are not ignored in its corresponding mini-batch. A value of 1 means the corresponding DTBB is a TP, and 0 means otherwise. 

\end{itemize}
First, we concatenate the respective vectors in $\mL$, $\mS$, and $\mD$, to obtain three vectors, $\vl$, $\vs$, and $\vd$. Second, we sort $\vl$ and $\vd$ by $\vs$. Third, we create two vectors $\vu = \vd$ and $\vv = \sim \vd$. The three vectors, $\vl$, $\vu$, and $\vv$ are used to simultaneously compute precision and recall values for every class.

Unlike sequential implementations, $\vl$, $\vu$, and $\vv$ correspond to DTBB's from all classes. To separate DTBB's by their classes, we transform $\vl$ to its one-hot encoding, $\mL \in \R^{z \times k}$, where $z$ is the total number of DTBB's that are not ignored, and k is the number of classes. For convenience, we transpose $\mL$ so that its shape is $k \times z$.

Since $\mL \in \R^{k \times z}$ and $\vu \in \R^z$ now have compatible shapes, we can obtain the class-specific TP mask, $\mU \in \R^{k \times z}$, by broadcasting $\vu$ over $\mL$ and applying the logical AND operator. If an element $\mU [ i ,j ]$ is equal to 1, it means that the $j$-th DTBB from the $i$-th class is a TP; otherwise, it means that the $j$-th DTBB in the 1-D slice $\mU [ i , : ]$ is not from the $i$-th class.

Similarly, we get the class-specific FP mask $\mV \in \R^{k \times z}$. If an element $\mV [ i ,j ]$ is equal to 1, it means that the $j$-th DTBB from the $i$-th class is an FP; otherwise, it means that the $j$-th DTBB in the 1-D slice $\mV [ i , : ]$ is not from the $i$-th class.

Having class-specific TP and FP masks, we can compute their cumulative sums. With their cumulative sums, we can finally calculate the precision and recall values for every class. Alg. \ref{algorithm_pr_computation} shows the pseudocode of our PR computation stage.
\begin{algorithm}[tb]

\caption{Parallel mAP Computation - PR Computation}

\begin{algorithmic}\label{algorithm_pr_computation}

\STATE {\bfseries Input:} $\sL, \sS, \sD, \hat{\vz}, k, \epsilon$ \# $\epsilon$ prevents division by zero

\STATE $\vl = \texttt{concatenate} ( \sL )$

\STATE $\vs = \texttt{concatenate} ( \sS )$

\STATE $\vd = \texttt{concatenate} ( \sD )$

\STATE Sort $\vl$ and $\vd$ by $\vs$

\STATE $\vu = \vd$

\STATE $\vv = \sim \vd$

\STATE $\mL = \texttt{tile} ( \texttt{expand\_dims} ( \vl , \texttt{axis} \texttt{=} 1 ) , ( 1 , k ) ) == \texttt{arange} ( k )$

\STATE $\mL = \texttt{transpose} ( \mL )$

\STATE $\mU = \mL \ \& \ \vu$

\STATE $\mV = \mL \ \& \ \vv$

\STATE $\mP = \texttt{cumsum} ( \mU , \texttt{axis} \texttt{=} 1 ) / \texttt{maximum} ( \texttt{cumsum} ( \mU , \texttt{axis} \texttt{=} 1 ) + \texttt{cumsum} ( \mV , \texttt{axis} \texttt{=} 1 ) , \epsilon )$

\STATE $\hat{\mZ} = \texttt{expand\_dims} ( \hat{z} , \texttt{axis} \texttt{=} 1 )$

\STATE $\mQ = \texttt{cumsum} ( \mU ,\texttt{axis} \texttt{=} 1 ) / \hat{\mZ}$

\STATE {\bfseries Output:} $\mP, \mQ$

\end{algorithmic}

\end{algorithm}
\subsection{mAP Calculation}
Having $\mP \in \R^{k \times z}$ and $\mQ \in \R^{k \times z}$, we can compute AP for each class in parallel. Again, we can either use every element in $\mP$ and $\mQ$ or use specific elements.
\subsubsection{Using Every Element}
If every element is used, our implementation calculates the exact areas under PR curves instead of the areas under the simplified curves, as mentioned in \ref{subsubsection_using_every_point}. The reason is that there lacks an efficient way to flatten the spikes on PR curves in a parallel fashion. Consequently, it is easier to calculate the exact mAP than the approximated mAP when done in parallel.

For each PR curve, we append a dummy point $( q = 0, p = 1 )$ at the start. The area under a PR curve is then equivalent to the sum of the trapezoidal areas under the curve. The area of the $i$-th trapezoid is defined in Equ. \ref{equation_trapezoid},
\begin{equation}\label{equation_trapezoid}
a_i = ( p_i + p_{i - 1} ) \cdot ( q_i - q_{i - 1} ) / 2
\end{equation}
where $i \in \{ 1, 2, \dotsc, z \}$, and $z$ is the total number of DTBB's that are not ignored. Alg. \ref{algorithm_map_calculation_using_every_element} shows the pseudocode of our mAP calculation stage using every element of $\mP$ and $\mQ$.
\begin{algorithm}[tb]

\caption{Parallel mAP Computation - mAP Calculation Using Every Element}

\begin{algorithmic}\label{algorithm_map_calculation_using_every_element}

\STATE {\bfseries Input:} $\mP, \mQ, k$

$\mP = \texttt{concatenate} ( \texttt{ones} ( ( k, 1 ) ) , \mP )$

$\mQ = \texttt{concatenate} ( \texttt{zeros} ( ( k, 1 ) ) , \mQ )$

$\text{mAP} = \texttt{mean} ( \texttt{sum} ( ( \mP [ : , 1 : ] + \mP [ : , : -1 ] ) \cdot ( \mQ [ : , 1 : ] - \mQ [ : , : -1 ] ) / 2 , 0 ) )$

\STATE {\bfseries Output:} mAP

\end{algorithmic}

\end{algorithm}
\subsubsection{Using Specific Elements}
If specific elements are used, we are again given a vector of recall levels $\hat{\vq}$. We expand $\mP$ and $\mQ$ such that they have compatible shapes with $\hat{\vq}$. We then use the rule in Sec. \ref{subsubsection_using_specific_points} to calculate the AP for every class at the same time. Alg. \ref{algorithm_map_calculation_using_specific_elements} shwos the pseudocode of mAP calculation stage using specific elements of $\mP$ and $\mQ$.
\begin{algorithm}[tb]

\caption{Parallel mAP Computation - mAP Calculation Using Specific Elements}

\begin{algorithmic}\label{algorithm_map_calculation_using_specific_elements}

\STATE {\bfseries Input:} $\mP, \mQ, \hat{\vq}$

\STATE $\tP = \texttt{tile} ( \texttt{expand\_dims} ( \mP , \texttt{axis} \texttt{=} 2 ) , ( 1 , 1 , \texttt{len} ( \hat{\vq} ) ) )$

\STATE $\tQ = \texttt{tile} ( \texttt{expand\_dims} ( \mQ , \texttt{axis} \texttt{=} 2 ) , ( 1 , 1 , \texttt{len} ( \hat{\vq} ) ) )$

\STATE $\text{mAP} = \texttt{mean} ( \texttt{amax} ( \tP \cdot ( \tQ \geq \hat{\vq} ) , \texttt{axis} \texttt{=} 1 ) )$

\STATE {\bfseries Output:} mAP

\end{algorithmic}

\end{algorithm}

\section{Experiments}\label{section_experiment}
ResNet-18 \citep{resnet} based CenterNet \citep{centernet} is used to produce DTBB's for the Pascal VOC 2007 test set, but it can be replaced with any other detector as long as DTBB's are generated in the required format. With the same DTBB's, our parallel implementation produces an mAP of 0.6658, whereas the sequential counterpart outputs an mAP of 0.6617. There is a 0.62\% difference between the two mAP values. We observe this slight inconsistency mainly comes from how operators used for determining TP's and FP's are optimized in the two underlying computational frameworks. Table \ref{tp_fp_diff} shows the number of TP's and FP's resulted from our way of computing mAP and the traditional way.
\begin{table}[tb]
\caption{Number of TP's and FP's from two implementations.}
\label{tp_fp_diff}
\begin{center}
\begin{tabular}{lllllll}
{\bf CLASS} &{\bf TP(P)} &{\bf TP(S)} &{\bf TP(D)} &{\bf FP(P)} &{\bf FP(S)} &{\bf FP(D)} \\
\\ \hline \\
aeroplane &239 &238 &0.42\% &12422 &12423 &0.01\% \\
bicycle &309 &307 &0.65\% &14675 &14677 &0.01\% \\
bird &387 &387 &0.00\% &20202 &20203 &0.00\% \\
boat &213 &211 &0.95\% &20193 &20191 &0.01\% \\
bottle &290 &288 &0.69\% &23588 &23589 &0.00\% \\
bus &192 &193 &0.52\% &12500 &12499 &0.01\% \\
car &1031 &1025 &0.59\% &41394 &41390 &0.01\% \\
cat &343 &342 &0.29\% &11875 &11876 &0.01\% \\
chair &613 &607 &0.99\% &61286 &61285 &0.00\% \\
cow &221 &221 &0.00\% &12613 &12611 &0.02\% \\
diningtable &180 &181 &0.55\% &17652 &17647 &0.03\% \\
dog &463 &461 &0.43\% &15401 &15401 &0.00\% \\
horse &330 &328 &0.61\% &12147 &12149 &0.02\% \\
motorbike &299 &297 &0.67\% &12258 &12259 &0.01\% \\
person &3770 &3750 &0.53\& &101778 &101796 &0.02\% \\
pottedplant &335 &334 &0.30\% &25441 &25444 &0.01\% \\
sheep &223 &219 &1.83\% &13615 &13615 &0.00\% \\
sofa &220 &220 &0.00\% &25808 &25796 &0.05\% \\
train &263 &262 &0.38\% &12069 &12070 &0.01\% \\
tvmonitor &272 &274 &0.73\% &15510 &15508 &0.01\% \\
\end{tabular}
\end{center}
\end{table}
\section{Conclusions}\label{section_conclusions}
Mean average precision has been the most indicative evaluation metric for object detection algorithms. However, implementations of computing it so far are inefficient and difficult to incorporate into training routines. As a result, almost no project tests mAP between each training epoch, although it is beneficial to see how a model improves at each step of the training process.

The problem with the traditional implementation of computing mAP is that it is performed sequentially, but machine learning procedures are desired to be executed in a parallel way. We find this problem needs to be attacked at various stages of the pipeline so that inputs and outputs are generated with fixed dimensions. Therefore, we propose a parallel and hardware acceleration compatible algorithm that delivers seamless integration with the main program.

Our experiments show that the resulting mAP value from our implementation only differs slightly from the one given by the original implementation. The difference is mainly caused by the low-level operators used by different computational frameworks and is insignificant in evaluating object detection algorithms.  
\newpage
\bibliography{iclr2020_conference}
\bibliographystyle{iclr2020_conference}

\end{document}